\documentclass[letterpaper, 10 pt, conference]{ieeeconf}
\IEEEoverridecommandlockouts                            
\usepackage[english]{babel}
\usepackage{amsmath} 
\usepackage{amssymb}  
\usepackage[dvips]{graphicx}
\usepackage{subfigure}
\usepackage{caption}
\usepackage{soul}
\usepackage{booktabs}
\usepackage{multirow}
\usepackage{dsfont}
\usepackage{amsfonts}
\usepackage{hyperref}
\usepackage{relsize}
\usepackage{tabularx}
\usepackage{algorithm,algpseudocode}
\usepackage{bm}
\usepackage{enumerate}
\usepackage{float}
\usepackage{siunitx}
\usepackage{mdframed}
\usepackage{authblk}
\usepackage{color}
\usepackage{xurl}
\usepackage{pifont}

\usepackage{booktabs}  
\usepackage{cite} 

\title{\LARGE \bf
Vision-based Relative Detection and Tracking\\ for Teams of Micro Aerial Vehicles
}

\author{
Rundong Ge$^1$$^*$, Moonyoung Lee$^1$$^*$, Vivek Radhakrishnan$^{1,2}$, Yang Zhou$^1$, Guanrui Li$^1$, and Giuseppe Loianno$^1$
\thanks{$^*$These authors contributed equally.}
\thanks{$^1$The authors are with the New York University, Tandon School of Engineering, Brooklyn, NY 11201, USA.
        {\tt\footnotesize email: \{rundong.ge, ml7617, vr2171, yangzhou, lguanrui, loiannog\}@nyu.edu.}}
\thanks{$^2$The author is with the Technology Innovation Institute, Abu Dhabi, UAE.
        {\tt\footnotesize email: vivek.radhakrishnan@tii.ae.}}
\thanks{This work was supported by the NSF CAREER Award 2145277, the NSF CPS Grant CNS-2121391, the Technology Innovation Institute, Qualcomm Research, Nokia, and NYU Wireless. Giuseppe Loianno serves as consultant for the Technology Innovation Institute. This arrangement has been reviewed and approved by the New York University in accordance with its policy on objectivity in research.}
}

\begin{document}

\maketitle
\thispagestyle{empty}
\pagestyle{empty}

\begin{abstract}
In this paper, we address the vision-based detection and tracking problems of multiple aerial vehicles using a single camera and Inertial Measurement Unit (IMU) as well as the corresponding perception consensus problem (i.e., uniqueness and identical IDs across all observing agents). We design several vision-based decentralized Bayesian multi-tracking filtering strategies to resolve the association between the incoming unsorted measurements obtained by a visual detector algorithm and the tracked agents. We compare their accuracy in different operating conditions as well as their scalability according to the number of agents in the team. This analysis provides useful insights about the most appropriate design choice for the given task.
We further show that the proposed perception and inference pipeline which includes a Deep Neural Network (DNN) as visual target detector is lightweight and capable of concurrently running control and planning with Size, Weight, and Power (SWaP) constrained robots on-board. Experimental results show the effective tracking of multiple drones in various challenging scenarios such as heavy occlusions.
\end{abstract}

\vspace{-5pt}
\section*{Supplementary Material}
\noindent \textbf{Video}: \url{https://youtu.be/mv5F7NLue6A}\\
\noindent \textbf{Code}: \url{https://github.com/arplaboratory/multi_robot_tracking}
\vspace{-10pt}
\section{Introduction}
\begin{figure}[!t]
    \centering
    \includegraphics[width=1\columnwidth]{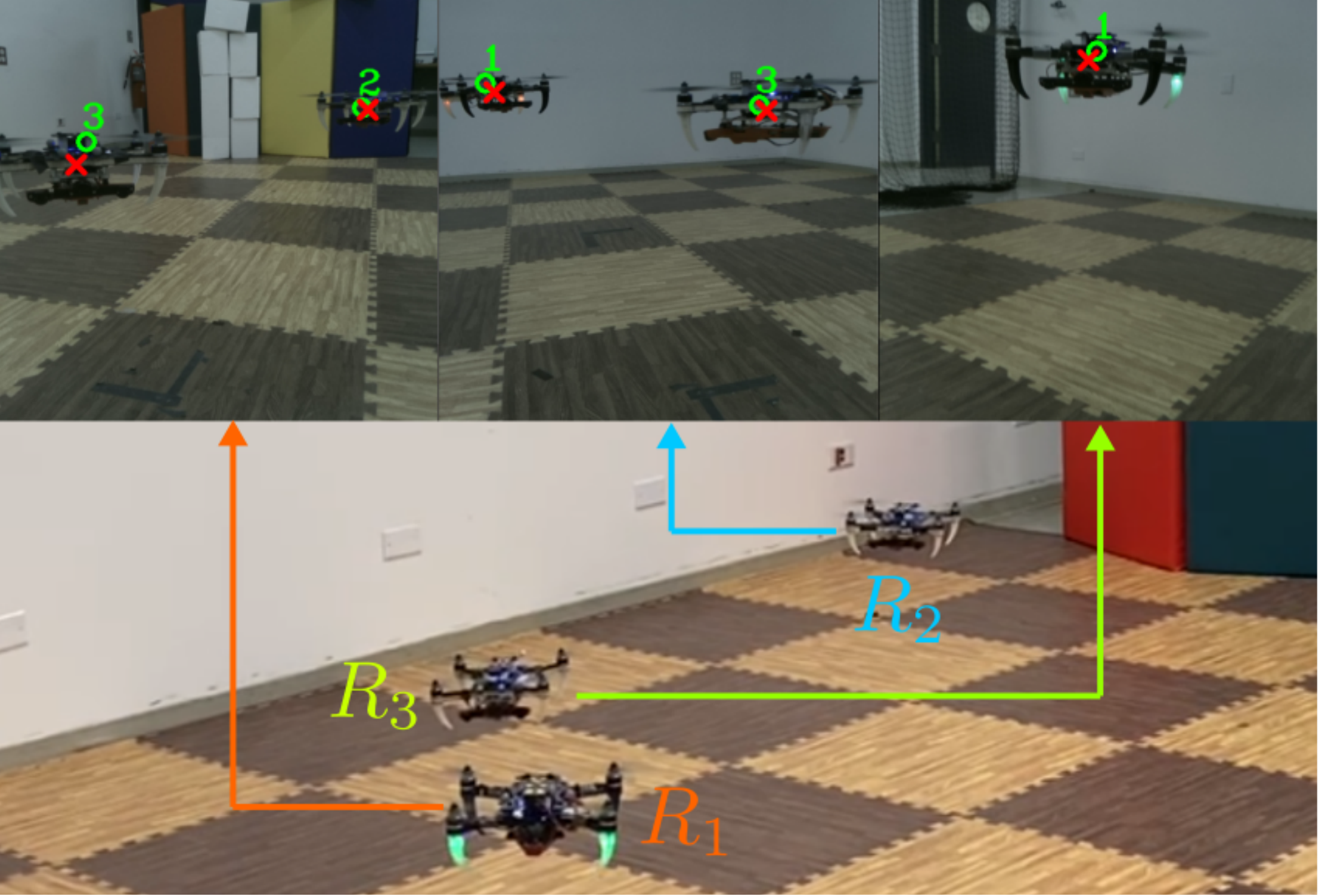}
    \caption{Multi-target tracking in each of the MAV's perspective. Tracked agents are in consensus among the three MAVs (green dots) using unsorted measurements (red crosses).}
    \label{fig:title_photo}
    \vspace{-20pt}
\end{figure}

Teams of Micro Aerial Vehicles (MAVs), often called swarms for large team sizes, are emerging as a disruptive on-demand technology to deploy distributed and intelligent autonomous systems~\cite{chung_survey_2018} for environment coverage, monitoring, situational awareness, transportation, and communication (e.g., creation of ad hoc remote networks). Application areas include but are not limited to agriculture, search and rescue, inspection, public safety (e.g., COVID-19 monitoring), warehouse management, and entertainment. By enabling cooperation among agents, multiple aerial vehicles offer additional flexibility, resilience, and robustness in several tasks compared to a single robot~\cite{Coppola2020}. However, several challenges remain in order to autonomously deploy them in real-world scenarios. To enable high-level autonomous decision-making policies, agents require on-board self-localization with respect to other robots~\cite{chung_survey_2018,Coppola2020} in a decentralized fashion with minimal communication. In this work, we address the vision-based decentralized detection and tracking for multiple MAVs with perception consensus among the agents as shown in Fig.~\ref{fig:title_photo}.
With a deep consideration for deploying MAVs constrained by Size, Weight, and Power (SWaP), we designed a lightweight perception pipeline that rely only on a single camera and Inertial Measurement Unit (IMU). These sensors have become popular due to low energy requirements of small-scale robots.
Early works on formation control~\cite{schiano_rigidity_maintenance_2017,schiano_dynamic_2018,zelazo2015decentralized} rely on an external motion capture system to detect and track the agents.
Our previous works~\cite{LoiannoIROS2016,WeinsteinRAL2018} address the on-board control and planning to generate collision-free formation trajectories in obstacle-free environments. However, the approach is not decentralized since the swarm framework relies on a ground station to resolve the relative detection and perception consensus. This latest work deploys team of MAVs without any central framework but instead only requires initial fleet configuration knowledge a priori.
The drone-to-drone relative localization has generally been addressed by additionally mounting artificial visual markers~\cite{Tron2016,dias2016board,Walter2019,Ahmad2021,Tang2019,Thakur2020} or 
employing a multi-modal sensor fusion approaches~\cite{Guo2017,PerezGrau2017, Xu2020}, including Ultra-Wide Band (UWB) which is the main focus in~\cite{Xu2020}. However, UWB technology still requires the installation of a specific system infrastructure.

Conversely, other works~\cite{Sapkota2016,Nguyen2020,Schilling2020,Pavliv2021} address the vision-based tracking problem. However,~\cite{Sapkota2016} does not focus on multiple vehicles. In~\cite{Nguyen2020}, the authors address the relative localization in a centralized manner. The approach relies on a powerful ground station to resolve localization and tracking. Moreover, only simulation results and evaluations on datasets are presented.  In~\cite{Schilling2020}, the tracking problem is resolved without explicit perception consensus among agents as well as bypassing the challenges related to deploying these approaches onto SWaP constrained robots. 
Furthermore, similar to~\cite{Tang2015} the authors employ only a specific filtering solution without comparative analysis. Finally, our previous work~\cite{Pavliv2021} only considered the detection and tracking problems in a centralized manner from a VR-headset. Conversely, in~\cite{Smith2019} the authors compare several tracking filtering solutions, but only provide simulation results and do not address the specificity of vision-based multi-tracking problem in different conditions neither the perception consensus. They also do not address the real-time deployment on SWaP constrained robots. Other works~\cite{Montijano2016,Cieslewski2018} achieve indirect relative localization by sharing a set of characteristic landmarks across the agents. Although these works relax the line-of-sight requirements, they require communication among the agents to share and store local maps. Recently, learning-based solutions for multi-target tracking have also surged in popularity~\cite{Ciaparrone2020}. However, these methods are computationally expensive to run on SWaP constrained robots.

The contributions of this paper are twofold.
First, we design several decentralized Bayesian vision-based multi-tracking filtering strategies to resolve the association problem between the incoming unsorted measurements obtained from a visual detector with perception consensus such that all observing drones agree on tracking targets' IDs. We compare their accuracy in different operating conditions as well as their scalability according to the number of agents in the team. Second, we show how the proposed setup including a Deep Neural Network (DNN) acting as visual target detector is able to run on-board in real-time on a small fleet of SWaP constraint MAVs concurrently with planning and control. The proposed pipeline utilizes only a minimalistic sensor suite composed of a single camera and IMU. To the best of our knowledge, this work shows the first vision-based detection and tracking for multiple MAVs in a decentralized manner where each agent is only equipped with a single camera and IMU while concurrently running planning and control in real-time. Overall, this approach can be deployed on--demand without relying on any external infrastructure or marker with the  potential  to  scale  to swarms of  aerial  robots. 

The paper is organized as follows. Section~\ref{sec:methodology} describes the proposed approach. Section~\ref{sec:comparative_analysis} analyzes the accuracy, computational complexity, and scalability of the methods. Section~\ref{sec:experiments} presents the experimental results, Section~\ref{sec:discussion} discusses the results, and Section~\ref{sec:conclusion} concludes the paper.

\section{Methodology}~\label{sec:methodology}
\vspace{-18pt}
\subsection{Preliminaries}\label{sec:preliminaries}
We consider a system of robots equipped with a camera and an IMU. Without loss of generality, we assume that for each robot, the camera and IMU frames are coincident with the robot frame. An external calibration procedure can compute the relative transformation between the frames. Our algorithm provides the state of each tracked agent in each robot or camera frame. We focus on the tracking problem without considering the relative pose problem (i.e., the estimated pose of each agent) that can be solved in parallel as in~\cite{Pavliv2021,Ge2021}.
In the following, we describe the multi-tracking procedure from an observing drone of a generic agent $i$. We design and analyze three Bayesian filters most representatives of several multi-target tracking filtering categories. Specifically, we design a Kalman Filter with the maximum likelihood of association (unimodal approach), a Joint Probabilistic Association filter (explicit computation of all possible association), and a Probability Hypothesis Density filter (random finite sets). These differ in the way each incoming $j^{th}$ measurement $\mathbf{z}_k^{j}\in Z_{k}$ obtained in the camera frame at time $k$ from a visual target detector (e.g., a DNN in our settings in Section~\ref{sec:experiments}) is associated with an existing tracked agent $\mathbf{x}_k^{i}$ with $i\in\{ 1\cdots n\}$. We denote the $j^{th}$ measurement associated with a tracked agent $\mathbf{x}_k^{i}$ as $\mathbf{z}_k^{i,j}\in Z_{k}$. In the filters, the IMU is used for the filter prediction whereas the visual measurements as update.

Each agent $i$ is tracked directly in the image plane using a 4-dimensional vector which contain the track positions and velocities along the $u$ and $v$ image axes
\begin{equation}
\mathbf{x}_{k}^{i}=\left[\begin{array}{llll}p_{u}^{i} & \dot{p}_{u}^{i} & p_{v}^{i} & \dot{p}_{v}^{i}
\end{array}\right]^{\top}.
\end{equation}
The relative motion between observed and target drones can then be represented by a stochastic nonlinear differential equation with a constant speed motion model
\begin{equation}
\label{eqn:motion_model}
\mathbf{x}_{k+1}^{i}=f\left(\mathbf{x}_{k}^{i},\mathbf{u}_{k},q\right)=\mathbf{A}_{k} \mathbf{x}_{k}^{i}+\mathbf{B}_{k}\left(\mathbf{x}_{k}^{i}\right) \mathbf{u}_{k} +\mathbf{Q}_{k},
\end{equation}
\begin{equation}
\mathbf{A}_{k}=\begin{bmatrix}
1 & \delta t & 0 & 0 \\
0 & 1 & 0 & 0 \\
0 & 0 & 1 & \delta t \\
0 & 0 & 0 & 1
\end{bmatrix}, ~ \mathbf{Q}_{k}=q\begin{bmatrix}
\frac{\delta t^{2}}{2} & 0 & 0 & 0 \\
0 & \delta t & 0 & 0 \\
0 & 0 & \frac{\delta t^{2}}{2} & 0 \\
0 & 0 & 0 & \delta
\end{bmatrix} 
\end{equation}
\begin{equation}
\mathbf{B}_{k}=\delta t\begin{bmatrix}
\frac{\left(p_{u}^{i}-c_{u}\right)\left(p_{v}^{i}-c_{v}\right)}{f} & -\frac{\left(p_{u}^{i}-c_{u}\right)^{2}}{f}-f & p_{v}^{i}-c_{v} \\
0 & 0 & 0 \\
f+\frac{\left(p_{v}^{i}-c_{v}\right)^{2}}{f} & -\frac{\left(p_{u}^{i}-c_{u}\right)\left(p_{v}^{i}-c_{v}\right)}{f} & -p_{u}^{i}+c_{u} \\
0 & 0 & 0
\end{bmatrix}
\end{equation}
where $\mathbf{u}_{k}$ is the angular velocity provided by the IMU in the robot frame of the observing agent, $\mathbf{A}_{k}$ is the state transition matrix, $\mathbf{B}_{k}$ is derived from the optical flow equation, and  $\mathbf{Q}_{k}$ is the process noise covariance matrix. Specifically, $\delta t$ is the sampling time at frame $k$, $q$ is the acceleration of the drones in $px/s^2$ assumed to be a Gaussian random variable, $(c_u,c_v)$ are the principal point coordinates, $f$ is the focal length. The reader can refer to~\cite{Pavliv2021} for more details of this model.

\vspace{-5pt}
\subsection{Multi-Target Tracking}

\subsubsection{Kalman Filter}
At every iteration of the Kalman filter, the algorithm includes a prediction and update steps.

\emph{Prediction step}: In this step, the robot computes the predicted state of each target using classic Kalman filter equations for the predicted mean and covariance based on the motion model equation  defined in eq.~(\ref{eqn:motion_model}).

\emph{Update step}: In the update step, before incorporating the measurement information for each tracked agent, it is necessary to associate each tracked agent $\mathbf{x}_{k}^{i}$ with a given measurement $j$ to apply the Kalman filter update equations obtained as updated mean and covariance respectively
\begin{equation}
\begin{aligned}
\bm{\mu}_{k \mid k}^{i} &= \bm{\mu}_{k \mid k-1}^{i}+\mathbf{K}_{k} \mathbf{y}_{k}^{i},
\\
\mathbf{P}_{k \mid k-1}^{i}&=\mathbf{F}_{k-1}\mathbf{P}_{k-1}^{i}\mathbf{F}_{k-1}^{\top}+\mathbf{Q}_{k-1},
\\\mathbf{F}_{k}&= \mathbf{A}_{k}+\frac{\partial \mathbf{B}_{k}}{\partial \mathbf{x}_{k}^{i}}\mathbf{u}_{k},
\label{eq:kalman_filter}
\end{aligned}
\end{equation}
where $\mathbf{y}_{k}^{i}$ = $\mathbf{z}_{k}^{i,j} - \mathbf{H}_{k}\bm{\mu}_{k \mid k-1}^{i}$ is the innovation term, $\mathbf{K}_{k}^{i}$ is the Kalman gain, $\bm{\mu}_{k \mid k-1}^{i}$ is the predicted mean of the state $\mathbf{x}_{k}^{i}$ at the prediction step, $\mathbf{H}_{k}$ is the measurement model defined in~\cite{Pavliv2021}. The association probability $p$ of the tracked agent $i$ with a measurement $j$ at time $k$ denoted as $\beta_{k}^{i,j}$ is obtained according to~\cite{Cordella2013} by selecting the maximum posterior distribution with respect to $\mathbf{x}_k^{i},\mathbf{z}_k^{j}$ pair as 
\begin{equation}
\begin{split}
&p\left(\beta_{k}^{i,j}\mid \mathbf{z}_k^{j},\mathbf{x}_k^{i}\right)\propto p\left(\mathbf{z}_k^{j}\mid\beta_{k}^{i,j},\mathbf{x}_k^{i}\right)p\left(\beta_{k}^{i,j}\right)=\\
&\mathcal{N}\left(\mathbf{z}_k^{j}-\bm{\mu}_{k \mid k-1}^{i},\mathbf{P}_{k \mid k-1}^i\right)p\left(\beta_{k}^{i,j}\right)
\end{split}\label{eq:prob_Kalman}
\end{equation}
where $\mathcal{N}$ is a normal distribution.
\subsubsection{Joint Probabilistic Data Association Filter}
The JPDAF is also divided into a prediction and update steps.

\emph{Prediction step}: In this step, the robot computes the predicted state in the same way as in the Kalman filter case.

\emph{Update step}: Similar to the Kalman filter, before incorporating the measurement information, the JPDAF explicitly resolves the association problem by computing all possible associations between the tracked agents and the incoming measurements.
These associations are represented in a matrix form as $\beta_{k}^{i}$. For each agent $i$, the update is performed using eq.~(\ref{eq:kalman_filter}), considering
\begin{equation}
\begin{aligned}
\mathbf{y}_{k}^{i} &=\sum_{j=1}^{Z_{k}} \beta_{k}^{i,j} \left(\mathbf{z}_{k}^{i,j} - \mathbf{H}_{k}\bm{\mu}_{k \mid k-1}^{i}\right),\\
\beta_{k}^{i}&=\sum_{\forall \chi} P\left\{\chi \mid Z_{k}\right\} \cdot \mathbf{I}(\chi)
\end{aligned}
\end{equation}
where $P\left\{\chi \mid Z_{k}\right\}$ the probability corresponding to each hypothesis matrix for the event $\chi$ and measurements $Z_{k}$, and $\mathbf{I}(\chi)$ the hypothesis matrix, $\beta_{k}^{i,j}$ is the $j^{th}$ column of $\beta_{k}^{i}$. While there are heuristics to reduce the computational burden associated with the explicit computation of all possible associations, such as proximity threshold, generating a hypothesis matrix for all possibilities can be computationally challenging once the number of tracked agents increases. the filter's computational complexity is discussed in Section~\ref{sec:comparative_analysis}.


\subsubsection{Gaussian Mixture PHD Filter} 
The GM-PHD is divided into a prediction and update step. It represents the measurement $\mathbf{z}_{k}^{j}$ and agent state $\mathbf{x}_{k}^{i}$ using Random Finite Sets (RFS) instead of explicitly associating all possible matches between the tracks and measurements. The tracked states can then be represented with density functions over the state space of targets, where the GM-PHD describes the first moment of distribution over the RFS. Each tracked agent state $\mathbf{x}_{k}^{i}$ can be described as a single intensity $v_{k}^{i}$ consisting of a weighted sum of Gaussian components in the form

\begin{equation}
v_{k}^{i}\left(\mathbf{x}_{k}^{i}\right)=\sum_{l=1}^{J_{k}} w_{k}^{l} \mathcal{N}\left(\mathbf{x}_{k}^{i} ;\bm{\mu}_{k}^{l}, \mathbf{P}_{k}^{l}\right),
\end{equation}
where the Gaussian components $\mathcal{N}$ for the state $\mathbf{x}_{k}^{i}$ is characterized by the weight ${w}_{k}^{l}$, mean $\bm{\mu}_{k}^{l}$, state covariance $\mathbf{P}_{k}^{l}$, with ${J}_{k}$ the number of tracked agents. Given the measurement and targets' previous states, the Gaussian components are propagated through the prediction and update steps. 

\emph{Prediction step}: Each agent state $i$ is still described as RFS
\begin{equation}
v_{k \mid k-1}^{i}\left(\mathbf{x}_{k}^{i}\right)=\sum_{l=1}^{J_{k}} w_{k \mid k-1}^{l} \mathcal{N}\left(\mathbf{x}_{k}^{i} ; \bm{\mu}_{k \mid k-1}^{l}, \mathbf{P}_{k \mid k-1}^{l}\right)+\gamma\left(\mathbf{x}_{k}^{i}\right),
\end{equation}
where $J_{k}$ is the number of tracked agents of the previous iteration and the corresponding Gaussian components adhere to the same motion model discussed in eq.~(\ref{eqn:motion_model}) with
\begin{equation}
\begin{aligned}
w_{k \mid k-1}^{l} &=p_{s} w_{k-1}^{l}, \\
\bm{\mu}_{k \mid k-1}^{l} &=\mathbf{A}_{k-1} \bm{\mu}_{k-1}^{l}+\mathbf{B}_{k-1} \mathbf{u}_{k-1} + \mathbf{Q}_{k-1},\\
\mathbf{P}_{k \mid k-1}^{l}&=\mathbf{F}_{k-1}\mathbf{P}_{k-1}^{l}\mathbf{F}_{k-1}^{\top}+\mathbf{Q}_{k-1},~\mathbf{F}_{k}= \mathbf{A}_{k}+\frac{\partial \mathbf{B}_{k}}{\partial \mathbf{x}_{k}^{i}}\mathbf{u}_{k}.
\end{aligned}
\end{equation}
Similarly to~\cite{Schilling2020}, we also assume an adaptive agent birth model $\gamma\left(\mathbf{x}_{k}^{i}\right)$ particular to the PHD filter in which new Gaussian components are characterized by ${w}_{\gamma}^{i}$, mean $\bm{\mu}_{\gamma}^{i}$, and covariance $\mathbf{P}_{\gamma}^{i}$. We do not have to account for association problem prior to new measurements, thus each agent's state can be set equal to the mean in the prediction step. We set the probability of survival ${p}_{s}$ of the target to $1$ since we assume the detected drones remain throughout the experiment.

\emph{Update step}: Each state RFS is being updated as 
\begin{equation}
\begin{aligned}
v_{k}\left(\mathbf{x}_{k}^{i}\right) &=\left(1-p_{d}\right) v_{k \mid k-1}\left(\mathbf{x}_{k}^{i}\right) \\
&+\sum_{j=1}^{Z_{k}}\sum_{l=1}^{J_{k}} w_{k}^{j,l}\left(\mathbf{z}_{k}^{j}\right) \mathcal{N}\left(\mathbf{x}_{k}^{i}; \bm{\mu}_{k \mid k}^{l}, \mathbf{P}_{k \mid k}^{l}\right),
\end{aligned}
\end{equation}
where ${p}_{d}$ is the probability of detection. The weight, mean, covariance, and Kalman gain updates are respectively
\begin{equation}
\begin{aligned}
w_{k}^{j,l} &=\frac{p_{d} w_{k \mid k-1}^{l} q_{k}^{l}\left(\mathbf{z}_{k}^{j}\right)}{\kappa_{k}+\sum_{l=1}^{J_{k \mid k-1}} p_{k}^{d} w_{k \mid k-1}^{l} q_{k}^{l}\left(\mathbf{z}_{k}^{j}\right)}, \\
\bm{\mu}_{k \mid k}^{j,l} &=\bm{\mu}_{k \mid k-1}^{l}+\mathbf{K}_{k}^{l}\left(\mathbf{z}_{k}^{j}-\mathbf{H}_{k} \bm{\mu}_{k \mid k-1}^{l}\right), \\
\mathbf{P}_{k \mid k}^{l}&=\left(\mathbf{I}-\mathbf{K}_{k}^{l} \mathbf{H}_{k}\right) \mathbf{P}_{k \mid k-1}^{l}, \\
\mathbf{K}_{k}^{l} &=\mathbf{P}_{k \mid k-1}^{l} \mathbf{H}_{k}^{\top}\left(\mathbf{H}_{k} \mathbf{P}_{k \mid k-1}^{l} \mathbf{H}_{k}^{\top}+\mathbf{R}_{k}\right)^{-1},
\end{aligned}
\end{equation}
where the clutter or the false positive term $\kappa_{k}$ is modeled as a random uniform distribution within the agent's field of view~\cite{PHAMNAMTRUNG2007}, $\mathbf{R}_{k}$ is the measurement noise covariance, and $q_{k}^{l}$ represents the probability of association between the $j^{th}$ measurement and $\bm{\mu}_{k \mid k-1}^{l}$ through a normal distribution 
\begin{equation}
\begin{aligned}
q_{k}^{l}\left(\mathbf{z}_{k}^{j}\right) &=\mathcal{N}\left(\mathbf{z}_{k}^{j}; \mathbf{H}_{k} \bm{\mu}_{k \mid k-1}^{l}, \mathbf{H}_{k} \mathbf{P}_{k \mid k-1}^{l} \mathbf{H}_{k}^{\top}+\mathbf{R}_{k}\right). \\
\end{aligned}
\end{equation}
After the weight and mean have been updated, we prune possible associations to keep computation manageable. To prune effectively, we discard components with weights less than the truncation threshold and by merging components with Mahalanobis distance less than a given merging threshold. The two thresholds are empirically determined to effectively discard false positives while robustly merging close associations during occlusion. Such pruning method yields only the agents with corresponding highest weights $w_k^{l}$, and the total remaining number of tracked agents $J_k$ is updated for the next iteration. Finally, the updated state of each track can be computed as a weighted sum of the associated mean as

\begin{equation}
\mathbf{x}_{k}=\frac{1}{\tilde{w}_{k}} \sum_{l=1}^{J_k} w_{k}^{l} \bm{\mu}_{k}^{l},~w_k^{l}\leftarrow{\max_{j}{w_k^{j,l}}\quad\forall j}
\end{equation}
where $\tilde{w}_{k}$ is the sum of the corresponding weights.
\vspace{-5pt}
\subsection{Perception Consensus}
Our framework does not utilize an external localization system. The multi-agent tracker handles association only in the local frame of each agent. Consequently, a challenge arises to have a global consensus to correctly tag uniqueness and identical tracked agents IDs among all observing drones. Without adopting a strong assumption of knowing the robots' global pose at all times as in~\cite{Dames2020}, we adopt a simple but efficient perception consensus module relying on minimum communication across the swarm. Each agent utilizes its own on-board localization system (e.g., VIO in Section~\ref{sec:experiments}), drone detection, and filter to solve the relative tracking problem.
\begin{algorithm}[!t]
\caption{Perception Consensus}
\hspace*{\algorithmicindent} \textbf{Input:}\text{ local 2D track[$N$] $T$, init 3D pose[$N$] $P$} \\
\hspace*{\algorithmicindent} \textbf{Output:}\text{ global 2D track[$N$] $ID$} 
\begin{algorithmic}[1]
\For{tracked targets $t$} 
\For{received init pose $i$}
\State $ProjP[i] = (u_p(i),v_p(i)) \gets Project2D(P[i])$ 
\State $\Delta p(t,i)\gets \|(T[t] - ProjP[i])\|_2$
\EndFor
\State $[val, index] \gets$  find minimum$(\Delta p(t,:))$
\State $ID[t] \gets  index$
\EndFor
\State \textbf{return} $ID$
\end{algorithmic}~\label{alg:consensus}
\vspace{-14pt}
\end{algorithm}
\begin{figure}[!b]
\vspace{-10pt}
    \centering
    \includegraphics[width=\linewidth]{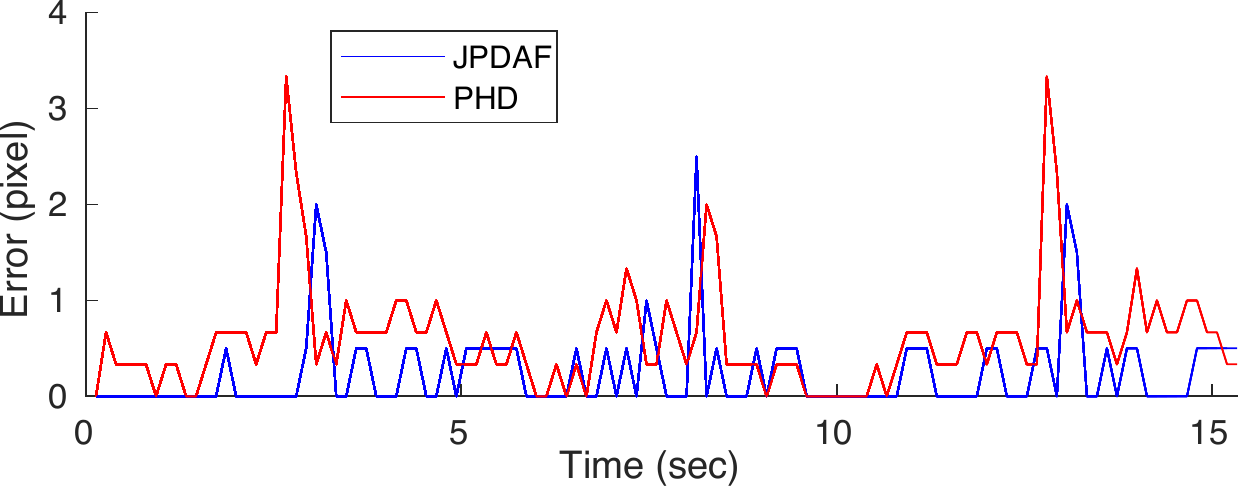}
    \caption{Average RMSE on tracking $3$ drones crossing paths.}
    \label{fig:error_over_time}
\end{figure}
\begin{figure}[!t]
    \centering
    \includegraphics[width=\linewidth]{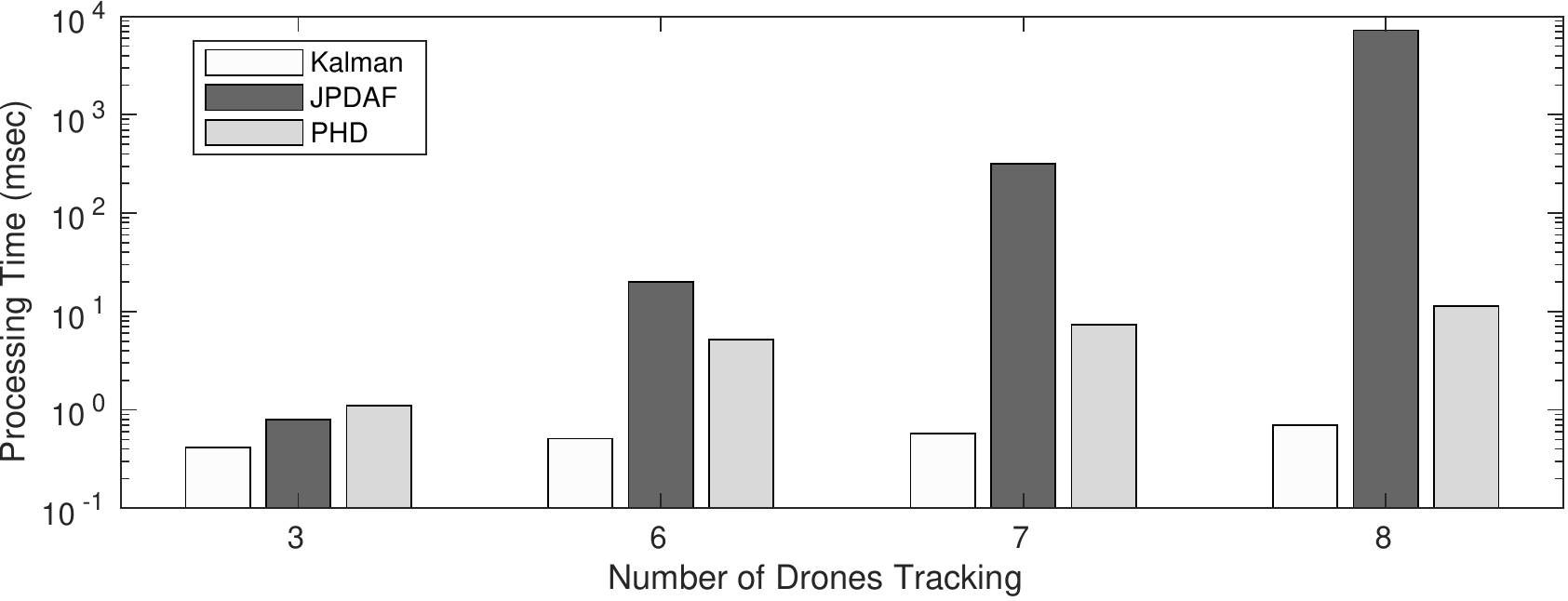}
    \caption{Computation time varying the number of drones.}
    \label{fig:computation_plot}
    \vspace{-20pt}
\end{figure}
A summary of the consensus method is shown in Algorithm \ref{alg:consensus}. It is assumed without loss of generality that each drone also has a unique ID assigned and is aware of its initial position in a global reference frame. 
The initial alignment between each local robot frame where the on-board localization is defined and the global frame can be computed at take-off using specific objects' visual features.
Subsequently, upon startup, the robots are within communication distance and field of view. Once initial tracking on each agent starts with one of the tracking filters, a one-time broadcast of the initial position of each drone and corresponding ID is performed.  Once shared, the 3D position of each drone is re-projected on the image plane, and the re-projection error for all possible tracked agents is computed. Each track ID in the filter of each observing drone will correspond to the drone ID with minimum re-projection error.
This procedure can be repeated to restore the consensus if tracking is re-initialized.
\section{Multi-Agent Tracking Comparative Analysis}~\label{sec:comparative_analysis}
We benchmark the performance of the Kalman, JPDAF and PHD multi-agent tracking algorithms. We discuss the trade-offs in terms of accuracy and computation complexity under varying operating conditions and number of agents. We simulate $3$ target drones crossing their trajectories in the image of an stationary observer drone. Given the unassociated and unsorted measurements, the filter estimates the 2D position tracking of the $3$ drones. In the presented tests, the average maximum speed for the $3$ drones is around $1$ m/s, but similar results hold for different speeds. Simulations are performed on an \text{Intel}\textsuperscript{\textregistered} i7 quad-core machine. 

\begin{figure*}[!t]
    \centering
    \subfigure [Evaluation for Kalman Filter against false positive, additive noise, and false negative. Increased intensity from top to bottom. The black crosses represent the unassociated raw measurement input. The tracking output of the Kalman filter over time is denoted in blue, red, green for the three targets.\vspace{-4pt}]	
    { 
    \includegraphics[width=1\linewidth]{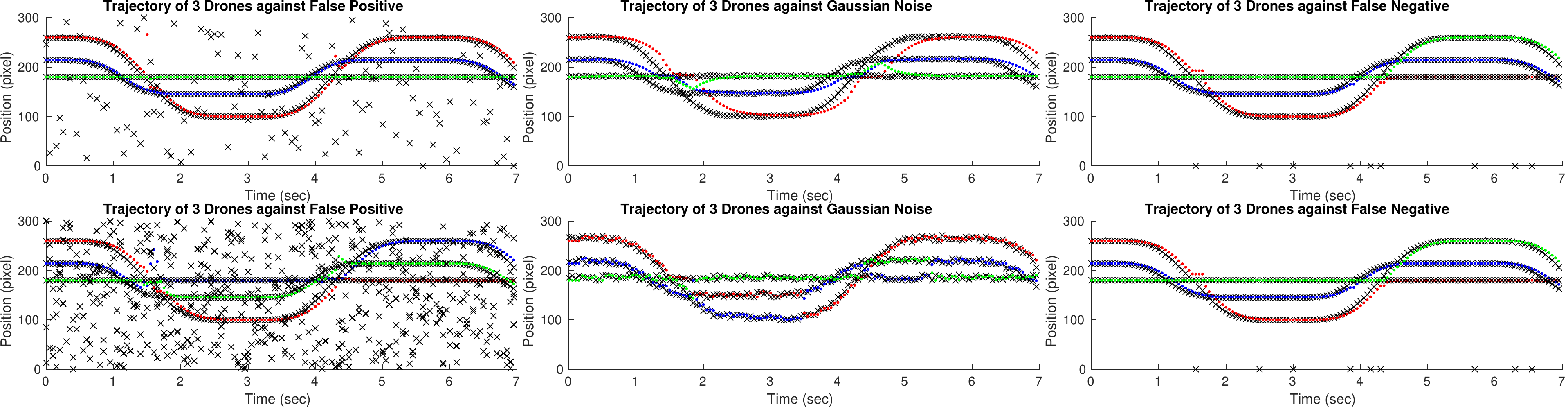}
    \label{fig:noisekalman}}
    \newline
    \subfigure [Evaluation for JPDAF filter against false positive, additive noise, and false negative. Increased intensity from top to bottom. The black crosses represent the unassociated raw measurement input. The tracking output of the JPDAF filter over time is denoted in blue, red, green for the three targets.\vspace{-4pt}]	
    { 
    \includegraphics[width=1\linewidth]{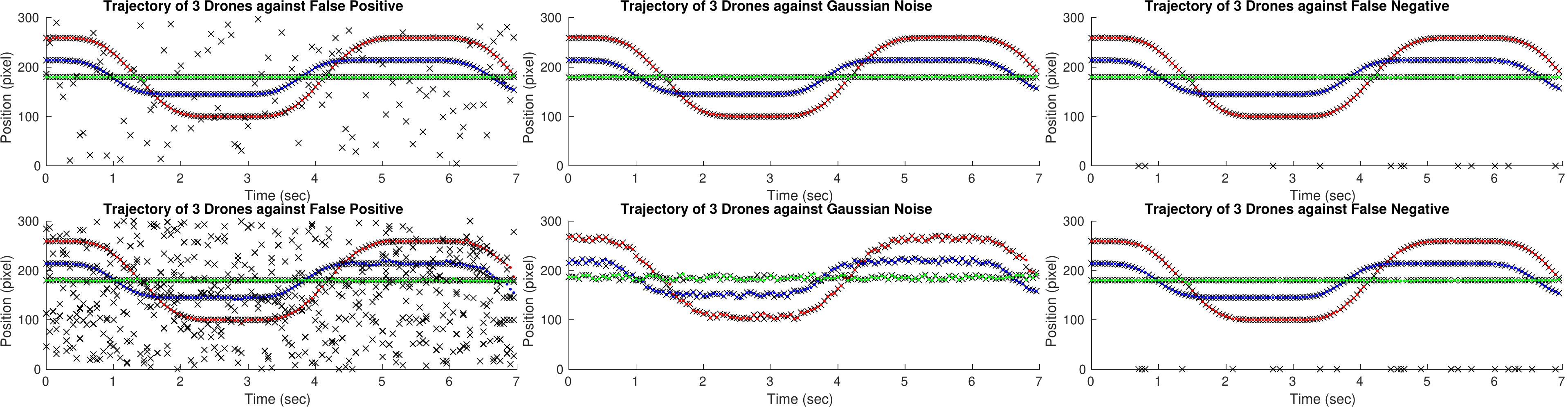}
    \label{fig:noisejpdaf}}
    \newline
    \subfigure [Evaluation for PHD filter against false positive, additive noise, and false negative. Increased intensity from top to bottom. The black crosses represent the unassociated raw measurement input. The tracking output of the PHD filter over time is denoted in blue, red, green for the three targets.\vspace{-4pt}]	
    { 
    \includegraphics[width=\linewidth]{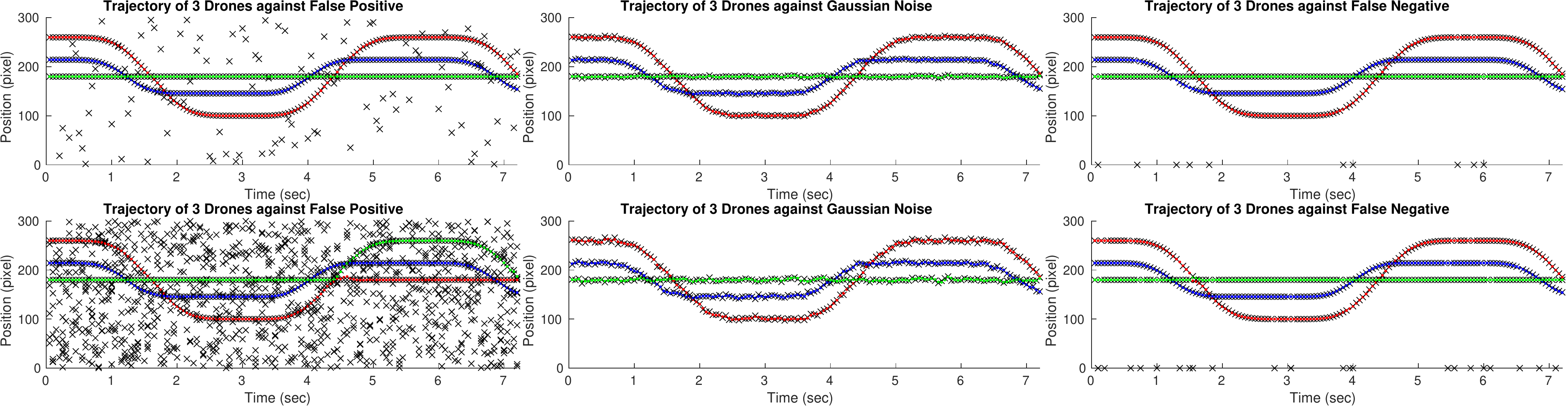}
    \label{fig:noisephd}}
    \newline
    \subfigure [Comparative analysis of the tracking errors as a function of noise intensities for each type of noise.\vspace{-4pt}]	
    {
    \includegraphics[width=0.33\linewidth]{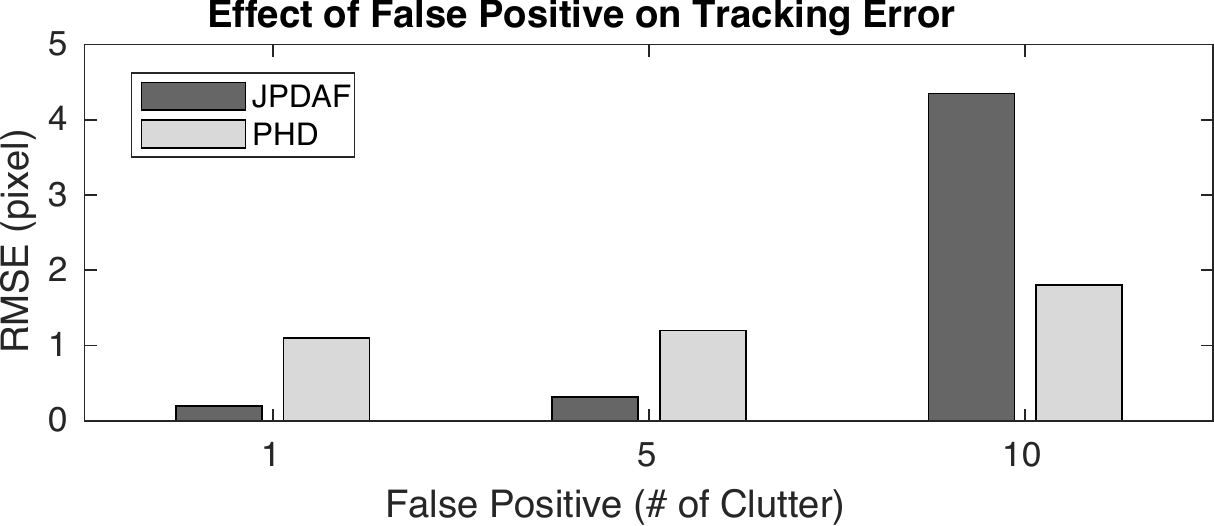}
    \includegraphics[width=0.33\linewidth]{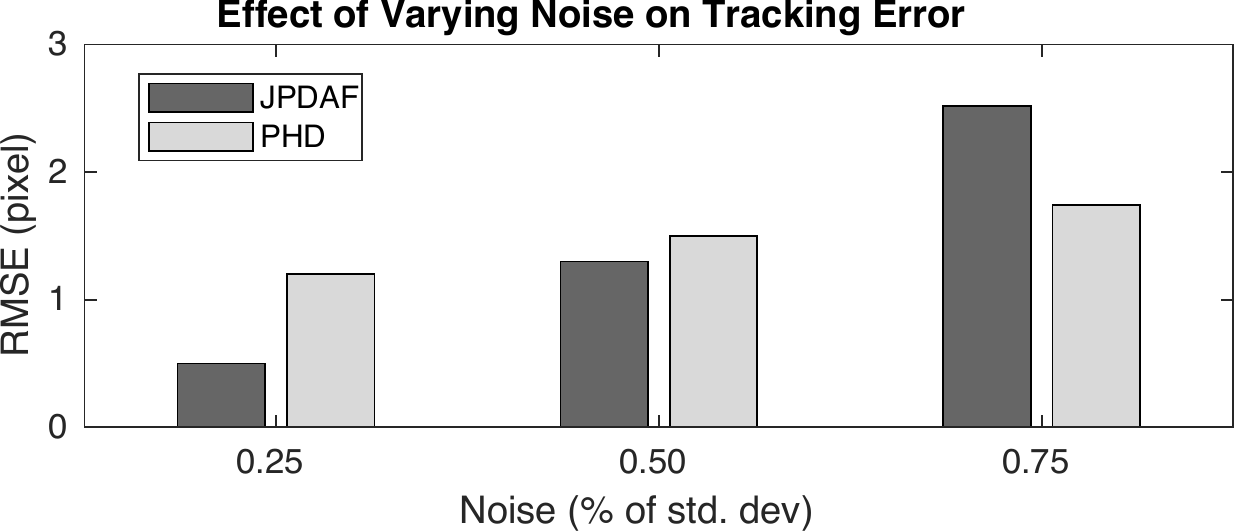}
    \includegraphics[width=0.33\linewidth]{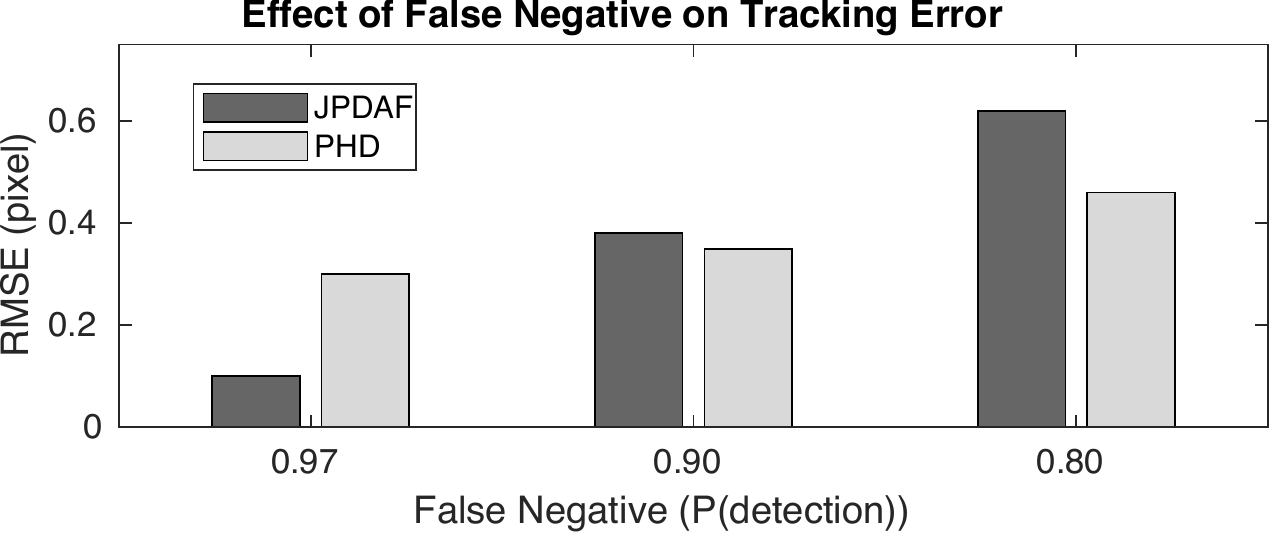}
    \label{fig:noiseRMSE}
    }
    \vspace{-10pt}
     \caption{
     Comparative performance analysis for different noise intensities. \vspace{-15pt}
    }
    \label{fig:noise}
\end{figure*}
\subsection{Accuracy vs. Computational Complexity}
In Fig.~\ref{fig:error_over_time}, the average tracking error (difference between tracked estimation and simulation ground truth of the $3$ drones) for both the JPDAF and GM-PHD filters is shown. We observe that the error is marginally higher for the GM-PHD comparing to the JPDAF filter. The GM-PHD tracking resulted in Root Mean Square Error (RMSE) of $0.735$ pixel while JPDAF tracking resulted in $0.302$ pixel. 
The RMSE of Kalman filter is $3.844$ pixels, which is much larger than those of the other two filters, so we did not add the RMSE of Kalman filter in Fig.~\ref{fig:error_over_time} to better highlight the comparison among the two best filters.
In Fig.~\ref{fig:computation_plot}, considering the aforementioned setup, we present the processing time of the three filters as a function of the number of tracked drones in log-scale. The Kalman filter's processing time increases linearly across the number of drones tracking. The JPDAF filter shows an exponential computation increase as the number of tracked drones increases, while the GM-PHD filter shows substantially lower computation complexity compared to the JPDAF filter. These results reflect the corresponding computational complexity of the two approaches, namely $O(N)$ for the Kalman filter, $O(N!)$ for the JPDAF~\cite{fortmann1980multi,Pavliv2021}, and $O(N^{2})$ for the PHD filter~\cite{Dames2020} with $N$ the number of tracked agents. The computation demand becomes untractable for the JPDAF while the number of tracked agents in each camera view increases, preventing its deployment for large swarms. Considering $8$ drones, the average computation time required at each iteration to resolve the association problem reaches $7.218$ s. On the other hand, the PHD filter is significantly faster, employing $0.011$ s, thus scaling better to track a large team. However, this comes at the price of a slighter lower accuracy compared to the JPDAF as shown in Fig.~\ref{fig:error_over_time}. Therefore, there is a trade-off between accuracy and computation since the PHD does not compute all associations between measurements and tracked agents as in the JPDAF.

\subsection{Noise Performance Evaluation}
We analyze the robustness of the proposed filters with respect to different types and intensities of measurement noise. We vary the number of false positives (number of measurement clutter from the environment, see Fig.~\ref{fig:noise} left column), injected Gaussian measurement noise (to emulate the noisy camera and IMU readings, see Fig.~\ref{fig:noise} center column), and false positives (probability of false detections, see Fig.~\ref{fig:noise} right column) to reflect real-world scenarios. The parameters of both filters have been set accordingly to match the various noise conditions. We show the tracking estimation (Figs.~\ref{fig:noisekalman},~\ref{fig:noisephd}, and~\ref{fig:noisejpdaf}) with intensities of false positive ($1$ and $10$ cases of clutter), Gaussian noise ($25\%$ and $75\%$ of the measurement noise covariance), and false negative ($0.97\%$ and $0.80\%$ probability of detection). We also show the tracking errors (Fig.~\ref{fig:noiseRMSE}). For each case, we vary the intensity and analyze how it affects the tracking performance.

The Kalman filter loses the agent tracking due to poor association as shown in Fig.~\ref{fig:noisekalman} (top row) or even switches tracked agents for large noise value as shown in Fig.~\ref{fig:noisekalman} (bottom row) compared to corresponding plots of other filters in Fig.s~\ref{fig:noisejpdaf} and~\ref{fig:noisephd}.
We can observe that in the evaluation against false positive (see Fig.~\ref{fig:noiseRMSE} left column), Gaussian noise (see Fig.~\ref{fig:noiseRMSE} center column), and false negative (see Fig.~\ref{fig:noiseRMSE} right column) the tracking RMSE is initially higher for the PHD compared to the JPDAF since the summation of Gaussian components affects the track. However, this weighted averaging effect proves to be more robust against high noise where the agent trajectories are crossing paths. 
The PHD maintains correct association during crossing whereas the JPDAF accrues larger error in intense clutter. For the false positive case of $10$ clutters (see Fig.~\ref{fig:noiseRMSE} left column), the PHD shows $1.81$ RMSE that is lower compared to $4.34$ of the JPDAF. For Gaussian Noise of $0.75\%$ of measurement noise covariance, PHD also shows $1.74$ pixels RMSE which is again lower compared to $2.52$ pixels RMSE of the JPDAF. For the false negative case of $0.8$ probability of detection, PHD again shows a lower RMSE of $0.45$ compared to $0.62$ that of JPDAF. For all other cases with lower noise intensities, we can conclude that the JPDAF shows better performances than the PHD filter. Similar results hold for different speeds.
We do not include the result of the Kalman filter in Fig.~\ref{fig:noiseRMSE} for ease of readability since the scale of the Kalman filter's RMSE is $10-100$ greater compared to those of the other two filters. The Kalman filter is $477.63$ pixels for the false positive with $10$ clutters while the JPDAF's RMSE is $4.35$ pixels. The high RMSE is due to tracking losses due to poor association in the Kalman filter compared to JPDAF and PHD filters. 
\section{Experimental Results}~\label{sec:experiments}
\vspace{-15pt}
\subsection{System Setup}
We report results from experiments with $3$ quadrotors conducted in an indoor flying space of $10\times6\times4~\si{m^3}$ at the Agile Robotics and Perception Lab (ARPL) lab at New York University. 
We employ custom small--scale aerial robots equipped with a Qualcomm\textsuperscript{\textregistered} $\text{Snapdragon}^{\text{TM}} \text{Flight}^{\text{TM}}$ Pro board and on-board VIO, planning, and control based on our previous work~\cite{LoiannoRAL2017}. The framework has been developped in ROS. 
Communication among drones is implemented using a synchronized multi-master network module~\cite{multimaster}.
Although alternative visual detectors would also work with the proposed tracking strategy, we empirically selected our detector that offered both robustness and inference speed as discussed in the next section. Once the tracking resolves the spatio-temporal association between target drones and the incoming measurements local to each robot, the perception consensus guarantees the uniqueness and identical IDs across all agents. 
\subsection{Visual Target Detector}\label{sec:detection}
We employ a DNN to to predict the 2D object centers as well as the regressed 2D bounding boxes from each observing agent's RGB front camera in real-time.
Our approach is inherited from CenterNet~\cite{zhou2019objects}. By directly regressing the objects' centers, CenterNet provides accurate and robust detection. CenterNet has a better tradeoff between speed and accuracy than YOLOv3~\cite{zhou2019objects}.
\begin{figure}[!t]
    \centering
    \includegraphics[width=1\columnwidth]{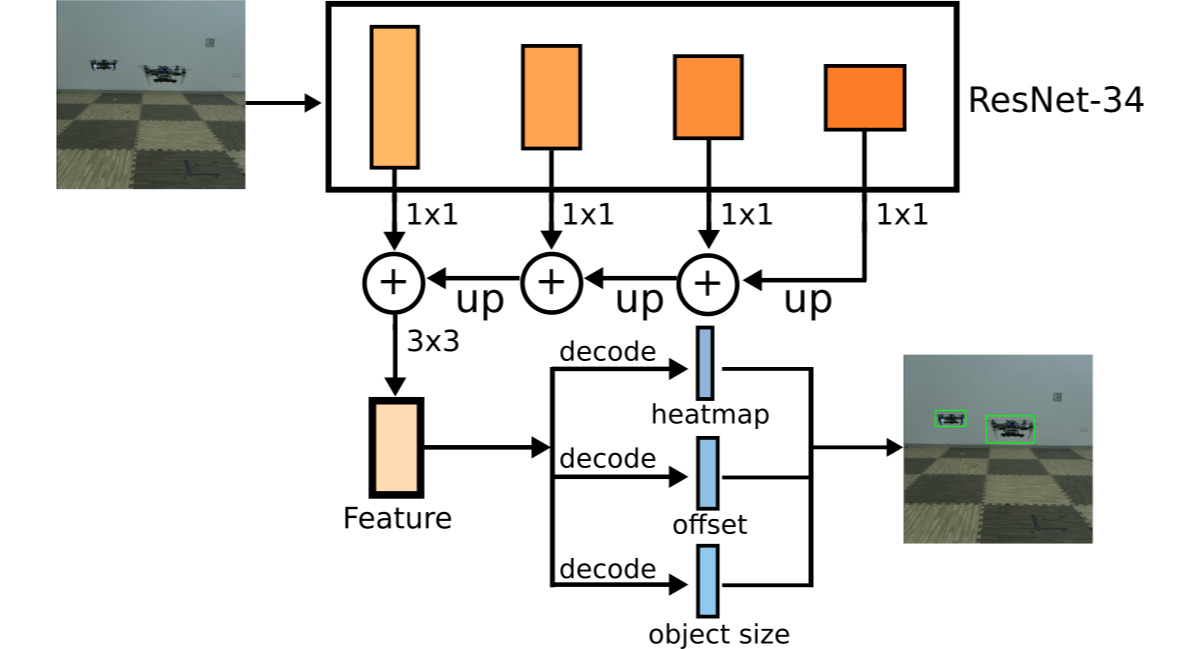}
    \caption{The proposed DNN architecture.}
    \label{fig:network_pipeline}
\vspace{-10pt}
\end{figure}
\begin{figure*}[!t]
    \centering
    \subfigure[drone 1 tracking viewpoint.]{
    \includegraphics[width=0.33\textwidth, trim=0 0cm 0 0, clip]{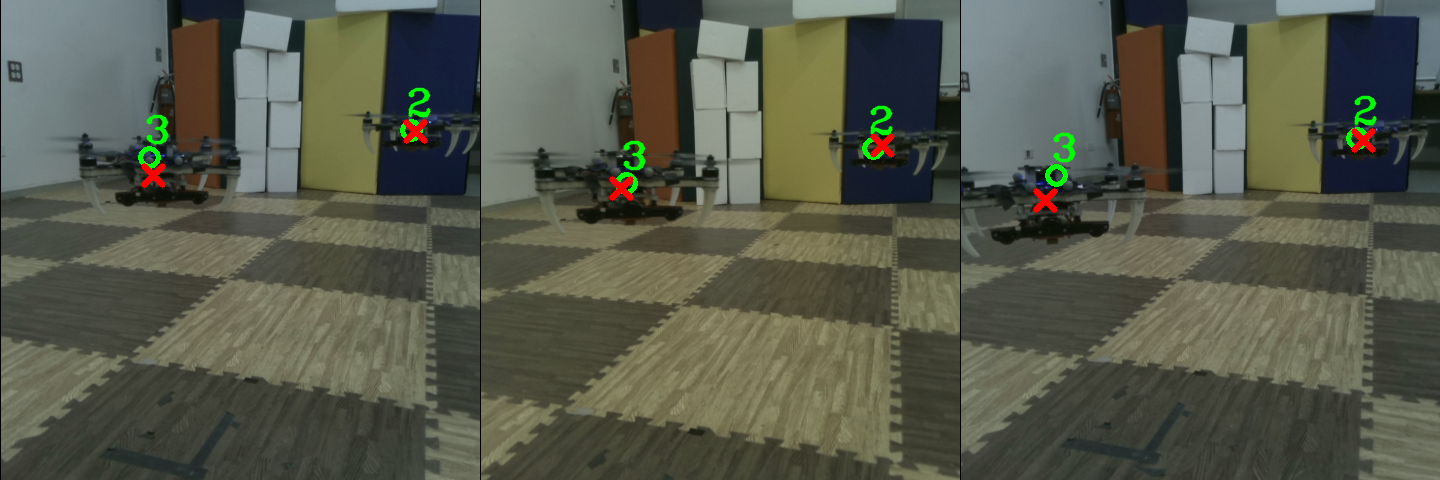}\hspace{-5pt}}
    \subfigure[drone 2 tracking viewpoint.]{
    \includegraphics[width=0.33\textwidth, trim=0 0cm 0 0, clip]{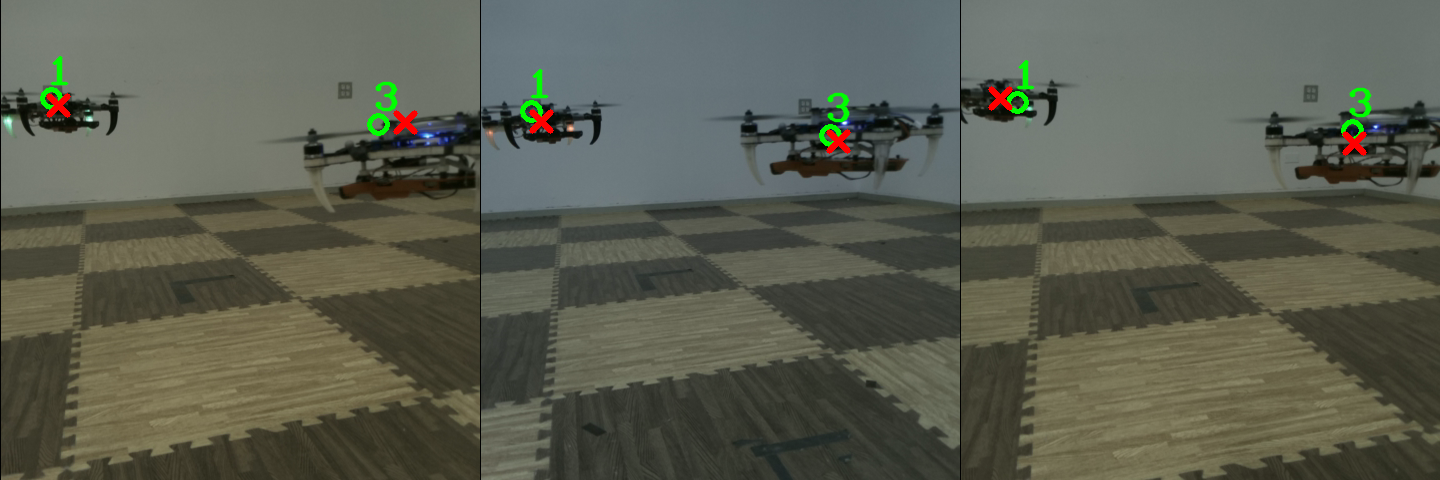}\hspace{-5pt}}
    \subfigure[drone 3 tracking viewpoint.]{
    \includegraphics[width=0.33\textwidth, trim=0 0cm 0 0, clip]{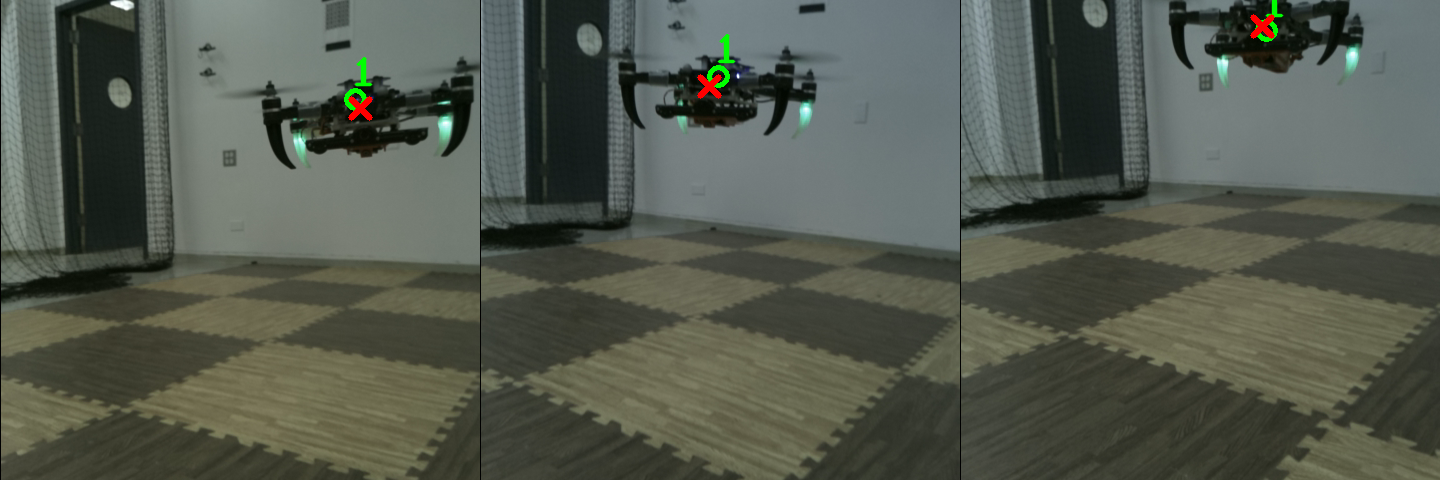}\hspace{-5pt}}
    \vspace{-10pt}
    \caption{Tracking performance on $3$ drones with current tracked agents (green) and measurements (red).}
    \label{fig:multi_robot}
    \vspace{-10pt}
\end{figure*}
\begin{figure*}[!t]
    \centering
    \includegraphics[width=1\textwidth, trim=0 5cm 0 0, clip]{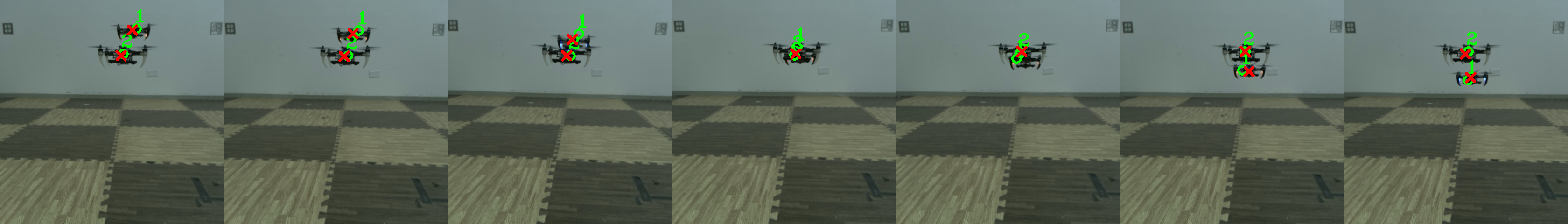}
    \includegraphics[width=1\textwidth, trim=0 5cm 0 0, clip]{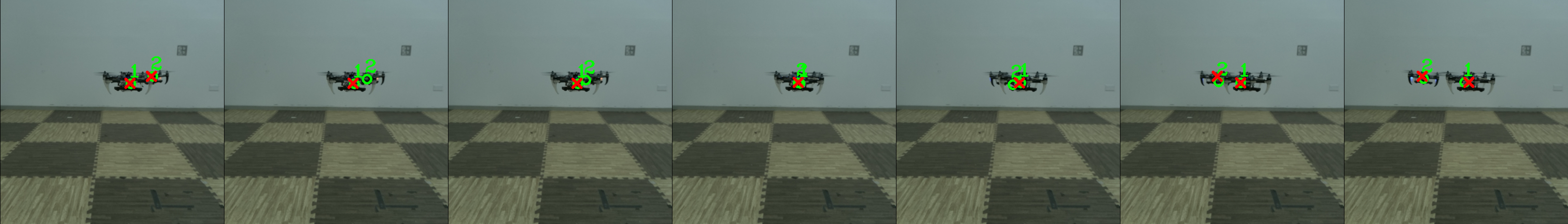}
    \caption{Tracking against occlusions. The two rows illustrate the time evolution of two sample challenging drone crossing scenarios with detections (red) and tracker prediction (green). Results show that our tracker is also robust to missed detections.}
    \label{fig:crossing}
    \vspace{-10pt}
\end{figure*}
The architecture is depicted in Fig.~\ref{fig:network_pipeline}. Let $I \in R^{W \times H \times 3}$ be the input image with width $W$ and height $H$. A ResNet-34 \cite{he2016deep} backbone augmented by three up-convolutional networks similar to~\cite{zhou2019objects} produces a center keypoint heatmap $\hat{Y} \in [0,1]^{\frac{W}{R} \times \frac{H}{R} \times C}$, where $R$ is the output stride (set to $4$) and $C$ is the number of classes (set to $1$). A local offset $\hat{O} \in \mathcal{R}^{\frac{W}{R} \times \frac{H}{R} \times 2}$ is regressed for each center point to recover the discretization error caused by the output stride. The object size $\hat{S} \in \mathcal{R}^{\frac{W}{R} \times \frac{H}{R} \times 2}$ is also regressed for each center point to obtain the bounding box. We train our model following~\cite{zhou2019objects}. 

We further adopt Qualcomm\textsuperscript{\textregistered} Snapdragon\textsuperscript{TM} Neural Processing Engine (SNPE) to deploy on-board our network. Our solution provides fast and efficient object detection at $7$ Hz with on-board GPU. We also noticed that CenterNet-based architectures are easier to deploy on edge devices due to the large number of operators compared to YOLOv3 representing another advantage of this approach.
\subsection{Results}
\vspace{-5pt}
To validate our proposed tracking system and compare the performance between Kalman, PHD, and JPDAF filters, we present various experiments where the $3$ drones move in each other's field of view. To further test the robustness of the proposed approaches to occlusions, we conduct several experiments where multiple drones are crossing their trajectories in the image of an observing agent. Our filters are able to estimate the full state of the vehicles at the IMU rate of $100$ Hz rather than waiting for a new detection which operates at $7$ Hz. This allows to speed up the inference by $14$ times and mitigates the problem of incorrect measurement association due to the slow $7$ Hz detection rate. Several experiments are also presented in the multimedia  material.

In the first experiment, $3$ drones move toward the same direction and do not cross trajectories while staying in each other's field of view. The tracking across the $3$ drones' viewpoints is shown in Fig.~\ref{fig:multi_robot}. The maximum relative speed between each target and the observing drone is $0.63$ m/s. The CPU usage is $7\%$. The raw images and all $3$ drones' position provided by Vicon 
are recorded for evaluation. The ground truth position of each drone for each image is obtained by projecting each drone's position on the image. The filter predicts the location of the observed drones while assigning globally consistent and unique IDs. In Table~\ref{table:result}, we report the RMSE between the ground-truth and estimated positions of the drones in all images for the Kalman, JPDAF, and PHD filters. This experiment has no occlusions or missed detections therefore JPDAF has lower RMSE than that of PHD, which is consistent with the simulation results in Fig.~\ref{fig:noiseRMSE} for low-medium noise settings. We also report the average computation time in Table~\ref{table:result}. These are consistent with simulation results in Section~\ref{sec:comparative_analysis} in Fig.~\ref{fig:computation_plot} since for less than 3 agents, $N=2$ in this case, the tracking computation cost of Kalman filter is lower compared to JPDAF which is more computationally efficient than PHD filter.

\begin{table}[!t]
\caption{Filters' accuracy and average computation time.}
\centering
 \resizebox{\linewidth}{!}{%
\begin{tabular}{c c c c c}
\toprule
\textbf{RMSE$\downarrow$} [pixels] & drone 1 & drone 2 & drone 3 & \textbf{Computation Time}$\downarrow$ (ms)\\ \midrule
Kalman        &    15.88    &   18.52      &    14.99   & 4.03     \\ \midrule
JPDAF     &   8.94      &    7.42     &     9.31   & 7.83 \\ \midrule
PHD       &   12.27      &   10.48      &    13.07  & 8.24   \\
\bottomrule
\end{tabular}
}
\label{table:result}
\vspace{-20pt}
\end{table}

\begin{table}[b]
    \vspace{-15pt}
    \caption{Discussion of three filters.}
    \centering
     \resizebox{0.7\linewidth}{!}{%
    \begin{tabular}{c c c c}
        \toprule 
        \textbf{Metric}$\uparrow$ & Kalman & JPDAF & PHD \\ \midrule
        Computation & ++++  & + & ++\\ \midrule
        Accuracy & + & ++++ & +++\\ \midrule
        Robustness & + & +++ & ++++ \\
        \bottomrule
    \end{tabular}
    }
    \label{tab:discussion}
\end{table}
Our second experiment shows that our system is robust to heavy occlusions by testing a hovering observing drone maintains tracks of two drones crossing their trajectories. This setting is challenging since the detection measurements are very close to each other or completely missed during the crossing. This can produce wrong association as shown in Fig.~\ref{fig:crossing}. We show that our filters consistently track the drones and maintain correct association for these cases except the Kalman filter. Consistent with what we observed in simulation results, the Kalman filter often incorrectly associates tracks among the agents, resulting in incomparably high RMSE. In contrast, the averaged RMSE of PHD filter across drones is $6.83$ pixels and $8.54$ pixels for the JPDAF filter.
These results are consistent with the simulation results in Fig.~\ref{fig:noiseRMSE}. We observe that the PHD filter outperforms the JPDAF filter due to the presence of false negative, higher noise and occlusions similar to the simulation results. This experiment has high occlusions since drone trajectories often cross causing missed detections as in Fig.~\ref{fig:crossing}.

\section{Discussion}~\label{sec:discussion}
Results in Sections~\ref{sec:comparative_analysis} and~\ref{sec:experiments} provide several insights on the filter's choice, according to three metrics, summarized in Table~\ref{tab:discussion}. From a computational perspective, the Kalman filter is the best solution. However, it struggles to be accurate and robust in presence of noise. Conversely, JPDAF and PHD are natural solutions for multi-target tracking providing increased robustness and accuracy. However, the use of the JPDAF is suggested for a small number of tracking agents ($N<3$). For larger teams, the computation becomes not manageable especially on SWaP constrained robots, therefore the PHD becomes the preferred solution in these cases.

\section{Conclusion}~\label{sec:conclusion} 
In this work, we presented several vision-based decentralized Bayesian filtering strategies for detection and tracking of multiple aerial vehicles and the corresponding perception consensus. We analyzed their trade-offs and infer the most suitable design based on the computation resources and scenarios. Finally, we showed how to transition this perception and inference pipeline onto SWaP constrained robots. The results show the effectiveness and robustness in challenging scenarios, including occlusions and missed detections with the potential to scale to swarms of robots.

Future works will investigate how to perform closed-loop formation control leveraging the tracking information. We envision extending the perception consensus to the absence of communication and incorporating this approach in a multi-modal framework, which can robustify the tracking and contribute to accurate multi-robot localization.




\bibliographystyle{IEEEtran} 
\bibliography{2022_IROS_Swarm_arxiv.bib}
\end{document}